%% file: main.tex
\pdfoutput=1

\documentclass[11pt]{article}

\usepackage[]{acl}

\usepackage{times}
\usepackage{latexsym}

\usepackage[T1]{fontenc}

\usepackage[utf8]{inputenc}
\usepackage{natbib}
\usepackage{microtype}
\usepackage{booktabs}
\usepackage{subcaption}
\usepackage{url}
\usepackage{comment}
\usepackage{todonotes}
\usepackage{color,soul}
\usepackage{graphicx}
\usepackage[normalem]{ulem}
\usepackage{algorithm}
\usepackage{algorithmic}
\usepackage{amsmath,amssymb}
\usepackage{longtable}
\usepackage{lscape}
\usepackage{enumitem}
\usepackage[normalem]{ulem}
\useunder{\uline}{\ul}{}

\title{Detecting the Role of an Entity in Harmful Memes: \\Techniques and Their Limitations}

\author{Rabindra Nath Nandi$^1$, Firoj Alam$^2$, Preslav Nakov$^2$\\
  $^1$BJIT Limited, Dhaka, Bangladesh \\
  $^2$Qatar Computing Research Institute, HBKU, Doha, Qatar \\
  \texttt{rabindra.nath@bjitgroup.com},
  \texttt{\{falam,pnakov\}@hbku.edu.qa}
  \\}

\begin{document}
\maketitle
\begin{abstract}
Harmful or abusive online content has been increasing over time, raising concerns for social media platforms, government agencies, and policymakers. Such harmful or abusive content can have major negative impact on society, e.g., cyberbullying can lead to suicides, rumors about COVID-19 can cause vaccine hesitance, promotion of fake cures for COVID-19 can cause health harms and deaths. The content that is posted and shared online can be textual, visual, or a combination of both, e.g., in a meme. Here, we describe our experiments in detecting the roles of the entities (hero, villain, victim) in harmful memes, which is part of the CONSTRAINT-2022 shared task, as well as our system for the task. We further provide a comparative analysis of different experimental settings (i.e., unimodal, multimodal, attention, and augmentation). For reproducibility, we make our experimental code publicly available.\footnote{\url{https://github.com/robi56/harmful_memes_block_fusion}}
\end{abstract}


\input{sections/introduction}
\input{sections/related_work}
\input{sections/task_dataset}
\input{sections/exp_setting}
\input{sections/results}
\input{sections/conclusions}

\bibliography{bib/all_bib,bib/bibliography}
\bibliographystyle{acl_natbib}

\end{document}

%% file: sections/introduction.tex
\section{Introduction}
\label{sec:intro}

Social media have become one of the main communication channels for sharing information online. Unfortunately, they have been abused by malicious actors to promote their agenda using manipulative content, thus continuously plaguing political events, and the public debate, e.g.,~regarding the ongoing COVID-19 infodemic \cite{alam-etal-2021-fighting-covid,10.1007/978-3-030-99739-7_52}. 
Such type of content includes harm and hostility \cite{brooke-2019-condescending,joksimovic-etal-2019-automated}, 
hate speech \citep{fortuna2018survey}, offensive language \cite{zampieri-etal-2019-predicting,SOLID}, abusive language \citep{mubarak2017abusive}, propaganda \cite{EMNLP19DaSanMartino,da2020survey}, cyberbullying \citep{van-hee-etal-2015-detection}, 
cyber-aggression \citep{kumar2018benchmarking}, and other kinds of harmful content \citep{pramanick-etal-2021-momenta-multimodal,Survey:2022:Harmful:Memes}.

The propagation of such content is often done by coordinated groups \cite{coordinated:communities:2022} using automated tools and targeting specific individuals, communities, and companies. There have been many research efforts to develop automated tools to detect such kind of content. 
Several recent surveys have highlighted these aspects, which include fake news \cite{zhou2020survey}, misinformation and disinformation \citep{alam2021survey,survey:media:2021,survey:stance:2022}, rumours \citep{bondielli2019survey}, propaganda \citep{da2020survey}, hate speech \citep{fortuna2018survey,Schmidt2017survey}, cyberbullying \citep{7920246}, offensive \citep{husain2021survey} and harmful content \citep{Survey:2022:Harmful:Memes}. 

The content shared on social media comes in different forms: textual, visual, or audio-visual. Among other social media content, recently, \textit{internet memes} became popular. Memes are defined as ``a group of digital items sharing common characteristics of content, form, or stance, which were created by associating them and were circulated, imitated, or transformed via the Internet by many users''~\citep{shifman2013memes}. Memes typically consist of images containing some text \citep{shifman2013memes,suryawanshi-etal-2020-multimodal,suryawanshi-etal-2020-dataset}. They are often shared for the purpose of having fun. However, memes can also be created and shared with bad intentions. This includes attacks on people based on characteristics such as ethnicity, race, sex, gender identity, disability, disease, nationality, and immigration status \cite{zannettou2018origins,kiela2020hateful}. There has been research effort to develop computational methods to detect such memes, such as detecting hateful memes \cite{kiela2020hateful}, propaganda \cite{dimitrov2021detecting}, offensiveness \cite{suryawanshi-etal-2020-multimodal}, sexist memes \cite{fersini2019detecting}, troll memes \cite{dravidiantrollmeme-eacl}, and generally harmful memes \cite{pramanick-etal-2021-momenta-multimodal,DISARM}. 

Harmful memes often target individuals, organizations, or social entities. \citet{pramanick-etal-2021-momenta-multimodal} developed a dataset where the annotation consists of (\emph{i})~whether a meme is harmful or not, and (\emph{ii})~whether it targets an individual, an organization, a community, or society. The CONSTRAINT-2022 shared task follows a similar line of research \cite{sharma2022report}. The entities in a meme are first identified and then the task asks participants to predict which entities are glorified, vilified, or victimized in the meme. The task is formulated as \textit{``Given a meme and an entity, determine the role of the entity in the meme: hero vs. villain vs. victim vs. other.''} More details are given in Section~\ref{sec:task_and_data}.

Memes are multimodal in nature, but the textual and the visual content in a meme are sometimes unrelated, which can make them hard to analyze for traditional multimodal approaches. Moreover, context (e.g., where the meme was posted) plays an important role for understanding its content. Another important factor is that since the text in the meme is overlaid on top of the image, the text needs to be extracted using OCR, which can result in errors that require additional manual post-editing \cite{dimitrov2021detecting}.

Here, we address a task about entity role labeling for harmful memes based on the dataset released in the CONSTRAINT-2022 shared task; see the task overview paper for more detail \cite{sharma2022report}. This task is different from traditional semantic role labeling in NLP \cite{palmer2010semantic}, where understanding \textit{who} did \textit{what} to \textit{whom}, \textit{when}, \textit{where}, and \emph{why} is typically addressed as a sequence labeling problem \cite{he-etal-2017-deep}. Recently, this has also been studied for visual content~\cite{sadhu2021visual}, i.e.,~situation recognition \cite{Yatskar_2016_CVPR,pratt2020grounded}, visual semantic role labeling \cite{gupta2015visual,silberer2018grounding,li2020cross}, and human-object interaction \cite{chao2015hico,chao2018learning}. 

To address the entity role labeling for a potentially harmful meme, we investigate textual, visual, and multimodal content using different pretrained models such as BERT \cite{text_bert}, VGG16 \cite{simonyan2014very}, and other vision--language models \cite{BlockFusion2019}. We further explore different textual data augmentation techniques and attention methods. For the shared task participation, we used only the image modality, which resulted in an underperforming system in the leaderboard.

Further studies using other modalities and approaches improved the performance of our system, but it is still lower (0.464 macro F1) than the best system (0.586). Yet, our investigation might be useful to understand which approaches are useful for detecting the role of an entity in harmful memes. 

Our contributions can be summarized as follows:
\begin{itemize} 
\item we addressed the problem both as sequence labeling and as classification;
\item we investigated different pretrained models for text and images;
\item we explored several combinations of multimodal models, as well as attention mechanisms, and various augmentation techniques.
\end{itemize}

The rest of the paper is organized as follows: Section~\ref{sec:related_work} presents previous work, Section~\ref{sec:task_and_data} describes the task and the dataset, Section~\ref{sec:classification} formulates our experiments, Section~\ref{sec:results} discusses the evaluation results. Finally, Section~\ref{sec:conclusion} concludes and points to possible directions for future work.

%% file: sections/related_work.tex
\section{Related Work}
\label{sec:related_work}

Below, we discuss previous work on semantic role labeling and harmful content detection, both in general and in a multimodal context.

\subsection{Semantic Role Labeling}

\paragraph{Textual semantic role labeling} has been widely studied in NLP, where the idea is to understand who did what to whom, when, where, and why. Traditionally, the task has been addressed using sequence labeling, e.g.,~\citet{fitzgerald2015semantic} used local and structured learning, experimenting with PropBank and FrameNet, and \citet{larionov2019semantic} investigated recent transformer models.

\paragraph{Visual semantic role labeling} has been explored for images and video. \citet{Yatskar_2016_CVPR} addressed situation recognition, and developed a large-scale dataset containing over 500 activities, 1,700 roles, 11,000 objects, 125,000 images, and 200,000 unique situations. The images were collected from Google and the authors addressed the task as a situation recognition problem. 
\citet{pratt2020grounded} developed a dataset for situation recognition consisting of 278,336 bounding-box groundings to the 11,538 entity classes.
\citet{gupta2015visual} developed a dataset of 16K examples in 10K images with actions and associated objects in the scene with different semantic roles for each action. 

\citet{yang2016grounded} worked on integrating language and vision with explicit and implicit roles. \citet{silberer2018grounding} learned frame–semantic representations of the images. \citet{sadhu2021visual} approached the same problem for video, developing a dataset of 29K 10-second movie clips, annotated with verbs and semantics roles for every two seconds of video content. 

\subsection{Harmful Content Detection in Memes}

There has been significant effort for identifying misinformation, disinformation, and malinformation online \cite{Schmidt2017survey,bondielli2019survey,zhou2020survey,da2020survey,alam2021survey,afridi2021multimodal,coordinated:communities:2022,10.1007/978-3-030-99739-7_52}. Most of these studies focused on textual and multimodal content. Compared to that, modeling the harmful aspects of memes has not received much attention.

Recent effort in this direction include categorizing hateful memes \cite{kiela2020hateful}, detecting antisemitism \cite{chandra2021subverting}, detecting the propagandistic techniques used in a meme \cite{dimitrov2021detecting}, detecting harmful memes and the target of the harm \cite{pramanick-etal-2021-momenta-multimodal}, identifying the protected categories that were attacked~\cite{zia-etal-2021-racist}, and identifying offensive content \cite{suryawanshi-etal-2020-multimodal}. Among these studies, the most notable low-level efforts that advanced research by providing high-quality datasets to experiment with include shared tasks such as the \emph{Hateful Memes Challenge} \cite{kiela2020hateful}, the SemEval-2021 shared task on detecting persuasion techniques in memes \cite{SemEval2021-6-Dimitrov}, and the troll meme classification task \cite{dravidiantrollmeme-eacl}.

\citet{chandra2021subverting} investigated antisemitism along with its types as a binary and a multi-class classification problem using  pretrained transformers and convolutional neural networks (CNNs) as modality-specific encoders along with various multimodal fusion strategies. \citet{dimitrov2021detecting} developed a dataset with 22 propaganda techniques and investigated the different state-of-the-art pretrained models, demonstrating that joint vision--language models performed better than unimodal ones. \citet{pramanick-etal-2021-momenta-multimodal} addressed two tasks: detecting harmful memes and identifying the social entities they target, using a multimodal model with local and global information. 

\citet{zia-etal-2021-racist} went one step further than a binary classification of hateful memes, focusing on a more fine-grained categorization based on the protected category that was being attacked (i.e.,~race, disability, religion, nationality, sex) and the type of attack (i.e.,~contempt, mocking, inferiority, slurs, exclusion, dehumanizing, inciting violence) using the dataset released in the WOAH 2020 Shared Task.\footnote{\url{http://github.com/facebookresearch/fine_grained_hateful_memes}} 
\citet{fersini2019detecting} studied sexist memes and investigated the textual cues using late fusion. They also developed a dataset of 800 misogynistic memes covering different manifestations of hatred against women (e.g.,~body shaming, stereotyping, objectification, and violence), collected from different social media~\citep{france_mys2021}.

\citet{kiela2021hateful} summarized the participating systems in the Hateful Memes Challenge, where the best systems fine-tuned unimodal and multimodal pre-training transformer models such as VisualBERT \citep{li2019visualbert} VL-BERT \citep{su2019vl}, UNITER \citep{chen2019uniter}, VILLA \citep{gan2020large}, and built ensembles on top of them.

The SemEval-2021 propaganda detection shared task \citep{SemEval2021-6-Dimitrov} focused on detecting the use of propaganda techniques in the meme, and the participants' systems showed that multimodal cues were very important. 

In the troll meme classification shared task \cite{dravidiantrollmeme-eacl}, the best system used ResNet152 and BERT with multimodal attention, and most systems used pretrained transformers for the text, CNNs for the images, and early fusion to combine the two modalities.

\paragraph{Combining modalities} causes several challenges, which arise due to representation issues (i.e.,~symbolic representation for language vs. signal representation for the visual modality), misalignment between the modalities, and fusion and transferring knowledge between the modalities. In order to address multimodal problems, a lot of effort has been paid to developing different fusion techniques such as (\emph{i})~{\em early fusion}, where low-level features from different modalities are learned, fused, and fed into a single prediction model \cite{jin2016novel,yang2018ti,zhang2019multi,spotfake,zhou2020safe,kang2020multi}, (\emph{ii})~{\em late fusion}, where unimodal decisions are fused with some mechanisms such as averaging and voting \cite{agrawal2017multimodal,qi2019exploiting}, and (\emph{iii})~{\em hybrid fusion}, where a subset of the learned features are passed to the final classifier (early fusion), and the remaining modalities are fed to the classifier later (late fusion) \cite{jin2017multimodal}. Here, we use early fusion and joint learning for fusion.

%% file: sections/task_dataset.tex
\section{Task and Dataset}
\label{sec:task_and_data}

Below, we describe the CONSTRAINT 2022 shared task and the corresponding dataset provided by the task organizers. More detail can be found in the shared task report \cite{sharma2022report}.

\subsection{Task}
\label{ssec:Task}
\begin{figure}
    \centering
    \includegraphics[width=0.47\textwidth]{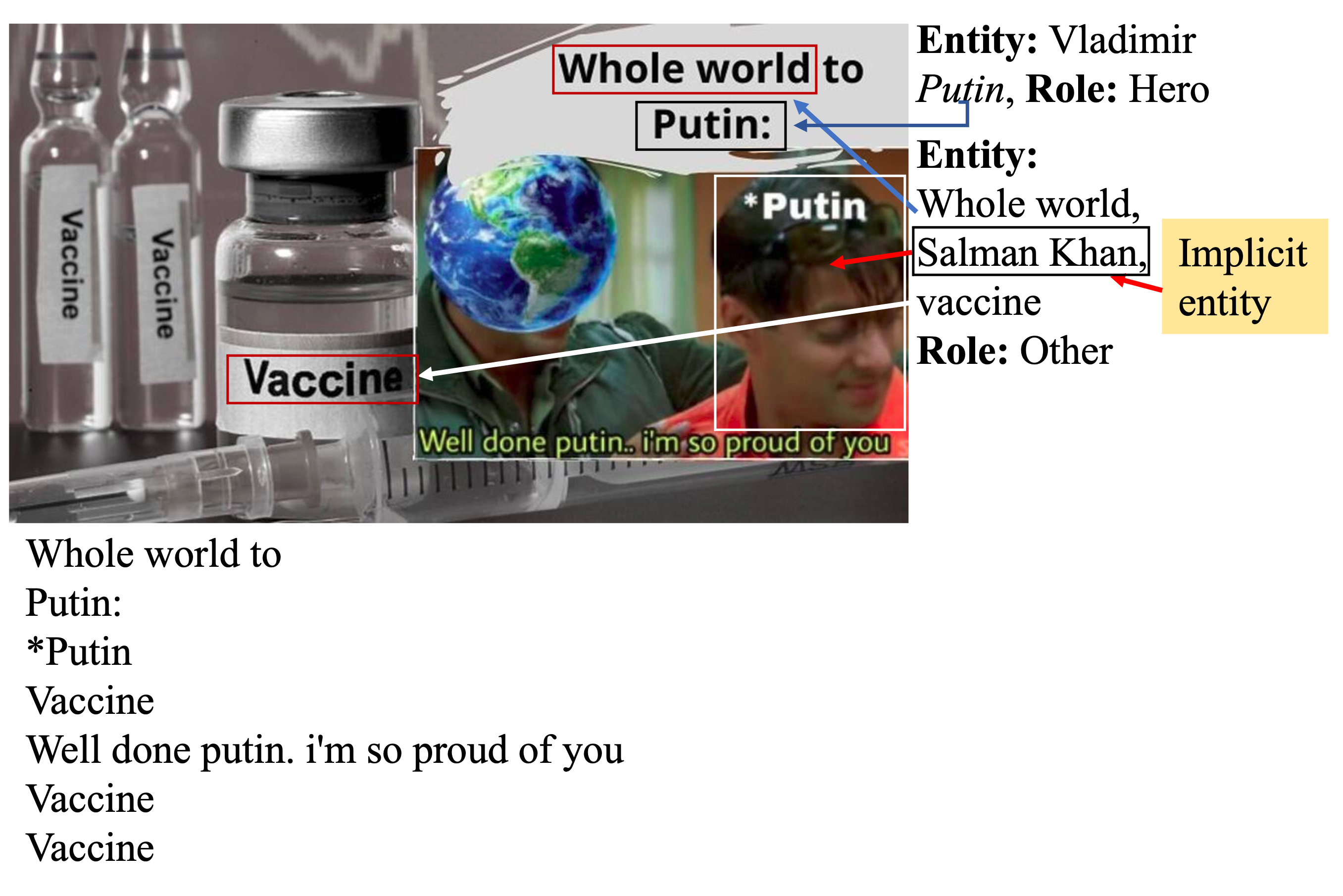}
    \caption{An example image showing the implicit (\emph{Salman Khan}) and the explicit entities (from a text perspective) and their roles.
    }
    \label{fig:task_example}
\end{figure}

The CONSTRAINT 2022 shared task asked participating systems to detect the role of the entities in the meme, given the meme and a list of these entities. Figure~\ref{fig:task_example} shows an example of an image with the extracted OCR text, implicit (image showing Salman Khan, who is not mentioned in the text), and explicit entities and their roles. The example illustrates various challenges: (\emph{i})~an implicit entity, (\emph{ii})~text extracted from the label of the vial, which has little connection to the overlaid written text, (\emph{iii})~unclear target entity in the meme (\emph{Vladimir Putin}). Such complexities are not common in the multimodal tasks we discussed above. The textual representation of the entities and their roles are different than for typical CoNLL-style semantic role labeling tasks \cite{carreras2005introduction}, which makes it more difficult to address the problem in the same formulation. 

By observing these challenges, we first attempted to address the problem in the same formulation: as a sequence labeling problem by converting the data to CoNLL format (see Section~\ref{sss:sequence_labeling}). Then, we further tried to address it as a classification task, i.e., predict the role of each entity in a given meme--entity pair. 

\subsection{Data}
\label{ssec:data}

We use the dataset provided for the CONSTRAINT 2022 shared task.
It contains harmful memes, OCR-extracted text from these memes, and manually annotated entities with four roles: \textit{hero}, \textit{villian}, \textit{victim}, and \textit{other}.
The datasets cover two domains: COVID-19 and US Politics. The COVID-19 domain consists of 2,700 training and 300 validation examples, while US Politics has 2,852 training and 350 validation examples. The test dataset combines examples from both domains, COVID-19 and US Politics, and has a total of 718 examples.

For the experiments, we combined the two domains, COVID-19 and US Politics, which resulted in 5,552 training and 650 validation examples.

The class distribution of the entity roles, aggregated over all memes, in the combined COVID-19 + US Politics dataset is highly imbalanced as shown in Table~\ref{tab:data_distribution}. We can see that overall the role of \emph{hero} represents only 2\%, and the role of \emph{victim} covers only 5\% of the entities. We can further see that the vast majority of the entities are labeled with the \emph{other} role. This skewed distribution adds additional complexity to the modeling task.

\begin{table}[]
\centering
\setlength{\tabcolsep}{2.0pt}
\scalebox{1.0}{
\begin{tabular}{@{}lrrrrrr@{}}
\toprule
\multicolumn{1}{c}{\textbf{Class label}} & \multicolumn{2}{c}{\textbf{Train}} & \multicolumn{2}{c}{\textbf{Val}} & \multicolumn{2}{c}{\textbf{Test}} \\ \midrule
\multicolumn{1}{c}{\textbf{}} & \multicolumn{1}{c}{\textbf{Count}} & \multicolumn{1}{c}{\textbf{\%}} & \multicolumn{1}{c}{\textbf{Count}} & \multicolumn{1}{c}{\textbf{\%}} & \multicolumn{1}{c}{\textbf{Count}} & \multicolumn{1}{c}{\textbf{\%}} \\ \midrule
Hero & 475 & 2 & 224 & 3 & 52 & 2 \\
Villain & 2,427 & 10 & 886 & 10 & 350 & 14 \\
Victim & 910 & 5 & 433 & 5 & 114 & 5 \\
Others & 13,702 & 83 & 6,937 & 82 & 1,917 & 79 \\ \midrule
Total & 17,514 &  & 8,480 &  & 2,433 & \\ \bottomrule
\end{tabular}
}
\caption{Distribution of the entity roles in the combined COVID-19 + US politics datasets.}
\label{tab:data_distribution}
\end{table}

%% file: sections/exp_setting.tex
\section{Experiments}
\label{sec:classification}

\paragraph{Settings:} We addressed the problem both as a sequence labeling and as a classification task. Below, we discuss each of them in detail. 

\paragraph{Evaluation measures:} In our experiments, we used accuracy, macro-average precision, recall, and F$_1$ score. The latter was the official evaluation measure for the shared task.

\subsection{Sequence Labeling}
\label{sss:sequence_labeling}
\begin{figure}
    \centering
    \includegraphics[width=0.47\textwidth]{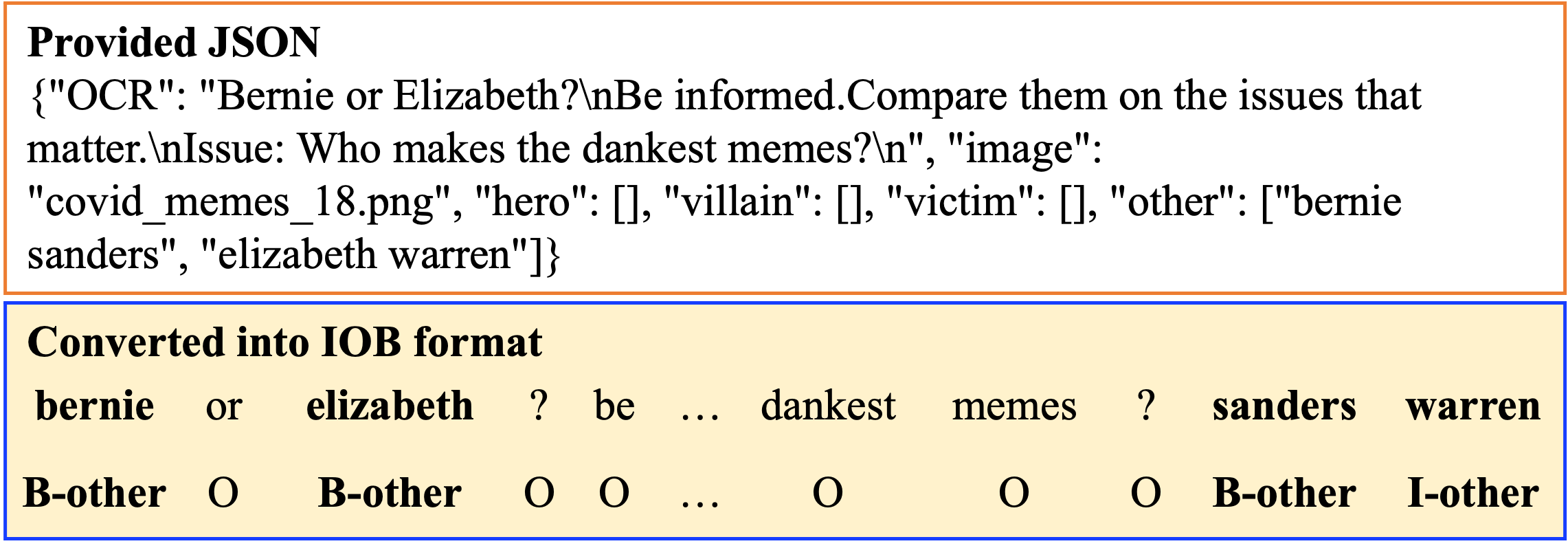}
    \caption{Example with text in BIO format.}
    \label{fig:data_iob_format}
\end{figure}

For the sequence labeling experiments, we first converted the OCR text and the entities to the CoNLL BIO-format. An example is shown in Figure~\ref{fig:data_iob_format}. To convert them, we matched the entities in the text and we assigned the same tag (role label) to the token in the text. For the implicit entity that is not in the text, we added them at the end of the text and we assigned them the annotated role; we labeled all other tokens with the O-tag. 

We trained the model using Conditional Random Fields (CRFs) \cite{lafferty2001conditional}, which has been widely used in earlier work. As features, we used part-of-speech tags, token length, tri-grams, presence of digits, use of special characters, token shape, w2vcluster, LDA topics, words present in a vocabulary list built on the training set, and in a name list, etc.\footnote{More details about the feature set can be found at \url{https://github.com/moejoe95/crf-vs-rnn-ner}} 
We ran two sets of experiments: (\emph{i})~using the same format, and (\emph{ii})~using only entities as shown in Figure~\ref{fig:data_iob_format}. 

\subsection{Classification}
\label{sss:classification}
For the classification experiments, we first converted the dataset into a classification problem. As it contains all examples with one or more entities, we reorganized the dataset so that an example contains an entity, OCR text, image, and entity role. Hence, the dataset size is now the same as the number of entity instances rather than memes. We ended up with 17,514 training examples, which is the number of training entities as shown in Table~\ref{tab:data_distribution}.

\begin{figure}
    \centering
    \includegraphics[width=0.48\textwidth]{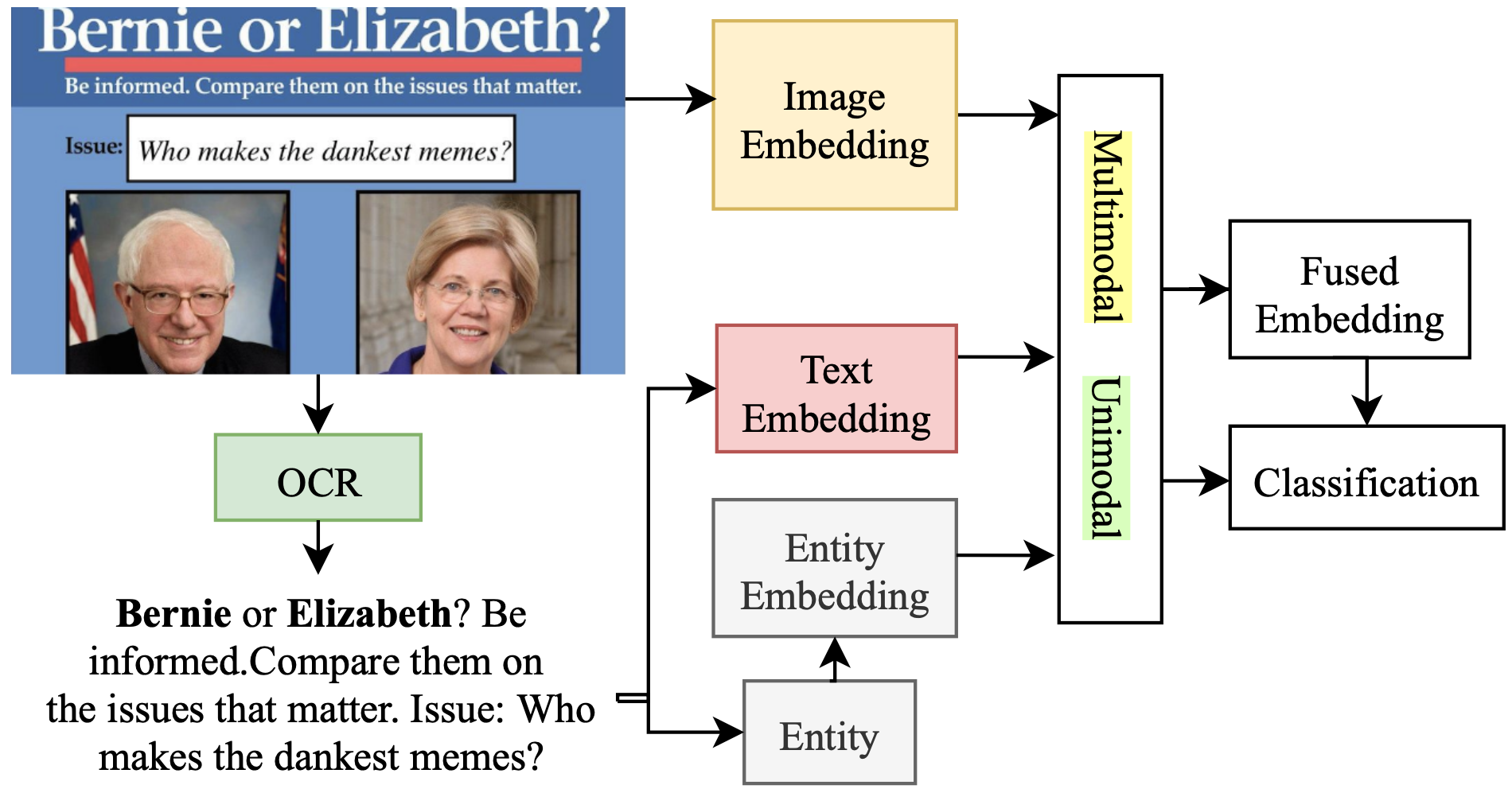}
    \caption{Diagram of our experimental pipeline.}
    \label{fig:experimental_pipeline}
\end{figure}

We then ran different unimodal and multimodal experiments: (\emph{i})~only text, (\emph{ii})~only meme, and (\emph{iii})~text and meme together. For each setting, we also ran several baseline experiments. We further ran advanced experiments such as adding attention to the network and text-based data augmentation. Figure~\ref{fig:experimental_pipeline} shows our experimental pipeline for this classification task. For the unimodal experiments, we used individual modalities, and we trained them using different pre-trained models. 

Note that for the text modality, we ran several combinations of fusion (e.g., text and entity) experiments. For the multimodal experiments, we combined embedding from both modalities, and we ran the classification on the fused embedding, as shown in Figure~\ref{fig:experimental_pipeline}.

\subsubsection{Text Modality}
For the text modality, we experimented using BERT \cite{text_bert} and XLM-RoBERTa \cite{liu2019roberta}. We performed ten reruns for each experiment using different random seeds, and then we picked the model that performed best on the development set. We used a batch size of 8, a learning rate of 2e-5, a maximum sequence length of 128, three epochs, and categorical cross-entropy as the loss function. We used the Transformer toolkit
to train the transformer-based models. 

Using the text-only modality, we also ran a different combination of experiments using the text and the entities, where we used  bilinear fusion to combine them. We discuss this fusion technique in more detail in Section~\ref{sssec:multimodal}. 

\subsubsection{Image Modality}
For our experiments using the image modality, we extract features from a pre-trained model, and then we trained an SVM classifier using these features. In particular, we extracted features from the penultimate layer of the EfficientNet-b1 (EffNet) model \cite{tan2019efficientnet}, which was trained using the ImageNet dataset. For training the model using the extracted features, we used SVM with its default parameter settings, with no further optimization of its hyper-parameter values. We chose EffNet as it was shown to achieve better performance for some social media image classification tasks \cite{alam2021medic,alam2021social}.

\subsubsection{Multimodal: Text and Image}
\label{sssec:multimodal}

For the multimodal experiments, we used the BLOCK Fusion~\cite{BlockFusion2019} approach, which was originally proposed for question answering (QA). Our motivation is that an entity can be seen like a question about the meme context, asking for its role as an answer. In a QA setting, there are three elements: (\emph{i})~a context (image or text), (\emph{ii})~a question, and (\emph{iii})~a list of answers. The goal is to select the right answer from the answer list. Similarly, we have four types of answers (i.e.,~roles). The task formation is that for an entity and a context (image or text), we need to determine the role of the entity in that context. 

BLOCK fusion is a multi-modal framework based on block-superdiagonal tensor decomposition, where tensor blocks are decomposed into blocks of smaller sizes, with the size characterized by a set of mode-$n$ ranks \cite{Lathauwer2008}. 
It is a bilinear model that takes two vectors $x^1 \in R^I$  and $x^2 \in R^J$ as input and then projects them to a $K$-dimensional space with tensor products: $y=\mathcal{T}\times x^1\times x^2$, 
where $y\in R^K$.  Each component of $y$ is a quadratic form of the inputs, $\forall k \in [1;K]$:

\begin{equation}
    y_k = \sum^I_{i=1}\sum^J_{j=1}\mathcal{T}_{ijk} x^1_i x^j_2 
\label{eq:block}
\end{equation}

BLOCK fusion can model bilinear interactions between groups of features, while limiting the complexity of the model, but keeping expressive high dimensional mono-model representations \cite{BlockFusion2019}. 
We used BLOCK fusion in different settings: (\emph{i})~for image and entity, (\emph{ii})~for text and entity, and (\emph{iii})~for text, image with entity.

\paragraph{Text and entity:} 
We extracted embedding representation for the entity and the text using a pretrained BERT model. We then fed both embedding representations into linear layers of 512 neurons each. The output of two linear layers is taken as input to the trainable block fusion network. Then, a regularization layer and linear layer are used before the final layer. 

\paragraph{Image and entity:} 
To build embedding representations for the image and the entity, we used a vision transformer (ViT) \cite{Dosovitskiy2021} and BERT pretrained models. The output of two different modalities was then used as input to the block fusion network. 

\paragraph{Image, text, and entity:}
In this setting, we first built embedding representations for the text and the image using a pretraind BERT and ViT models, respectively. Then, we concatenated these representations (text + image) and we passed them to a linear layer with 512 neurons. We then extracted embedding representation for the target entity using the pretraind BERT model. Afterwards, we merged the text + image and the entity representations and we fed them into the fusion layer. In this way, we combined the image and the text representations as a unified context, aiming to predict the role of the target entity in this context. 

In all the experiments, we uses a learning rate of $1e^{-6}$, a batch size of 8, and a maximum length of the text of 512.

\subsubsection{Additional Experiments}
We ran two additional sets of experiments using attention mechanism and augmentation, as using such approaches has been shown to help in many natural language processing (NLP) tasks. 

\paragraph{Attention:} 
In the entity + image block fusion network, we used block fusion to merge the entity and the image representations. Instead of using the image representation directly, we used attention mechanism on the image and then we fed the attended features along with the entity representation into the entity + image block. To compute the attention, we used the PyTorchNLP library.\footnote{\url{http://github.com/PetrochukM/PyTorch-NLP}}
In a similar fashion, we applied the attention mechanism to the text and to the combined text + image representation.

\paragraph{Augmentation:} 
Text data augmentation has recently gained a lot of popularity as a way to address data scarceness and class imbalance \cite{feng2021survey}. We used three types of text augmentation techniques to balance the distribution of the different class: (\emph{i})~synonym augmentation using WordNet, (\emph{ii})~word substitution using BERT, and (\emph{iii})~a combination thereof. In our experiments, we used the NLPAug data augmentation package.\footnote{\url{https://github.com/makcedward/nlpaug}} 
Note that we applied six times augmentation for the \emph{hero} class, twice for the \emph{villain} class, and three times for the \emph{victim} class. These numbers were empirically set and require further investigation in future work. 

%% file: sections/results.tex
\section{Results and Discussion}
\label{sec:results}

Below, we first discuss our sequence labeling and classification experiments. We then perform some analysis, and finally, we put our results in a broader perspective in the context of the shared task.

\subsection{Sequence Labeling Results}

Table~\ref{tab:sequence_classification_results} shows the evaluation results on the test set for our sequence labeling reformulation of the problem. We performed two experiments: one where we used as input the entire meme text (i.e.,~all tokens), and another one where we used the concatenation of the target entities only. We can see that the latter performed marginally better, but overall the macro-F1 score is quite low in both cases.

\begin{table}[t]
\centering
\setlength{\tabcolsep}{3.0pt}
\scalebox{0.95}{
\begin{tabular}{@{}lrrrr@{}}
\toprule
\multicolumn{1}{c}{\textbf{Exp.}} & \multicolumn{1}{c}{\textbf{Acc}} & \multicolumn{1}{c}{\textbf{P}} & \multicolumn{1}{c}{\textbf{R}} & \multicolumn{1}{c}{\textbf{F1}} \\ \midrule
 All tokens & 0.51 & 0.32 & 0.21 & 0.24 \\
 Only entities & 0.77 & 0.40 & 0.27 & 0.25 \\ 
\bottomrule
\end{tabular} 
}
\caption{Evaluation results on the test set for the sequence labeling reformulation of the problem.}
\label{tab:sequence_classification_results}
\end{table}

\subsection{Classification Results}

Table \ref{tab:classification_results} shows the evaluation results on the test set for our classification reformulation of the problem. We computed the \textit{majority class} baseline (row 0), which always predicts the most frequent label in the training set. Due to time limitations, our official submission used the image modality only, which resulted in a very low macro-F1 score of 0.23, as shown in row 1. 
For our text modality experiments, we used the meme text and the entities. We experimented with BERT and XLM-RoBERTa, obtaining better results using the former. Using the BLOCK fusion technique on unimodal (text + entity) and multimodality (text + image + entity) yielded sizable improvements. The combination of image + text (rows 6 and 9) did not yield much better results compared to using text only (row 4).
Next, we added attention on top of block fusion, which improved the performance, but there was no much difference between the different combinations (rows 7--9).
Considering only the text and the entity, we observe an improvement using text augmentation. Among the different augmentation techniques, there was no performance difference between WordNet and BERT, and combining them yielded worse results.

\begin{table}[h]
\centering
\setlength{\tabcolsep}{2.0pt}
\scalebox{0.9}{
\begin{tabular}{@{}llrrrr@{}}
\toprule
 & \multicolumn{1}{c}{\textbf{Exp.}} & \multicolumn{1}{c}{\textbf{Acc}} & \multicolumn{1}{c}{\textbf{P}} & \multicolumn{1}{c}{\textbf{R}} & \multicolumn{1}{c}{\textbf{F1}} \\ \midrule
\multicolumn{6}{c}{\textbf{Baseline}} \\ \midrule
0 & Majority & 0.79	& 0.20 & 0.25 & 0.22 \\ \midrule
\multicolumn{6}{c}{\textbf{Image modality}} \\ \midrule
\it 1 & \it EffNet feat + SVM & \it 0.72 & \it 0.24 & \it 0.25 & \it 0.23 \\ \midrule
\multicolumn{6}{c}{\textbf{Text modality}} \\ \midrule
2 & BERT & 0.76 & 0.42 & 0.36 & 0.37 \\
3 & XLM-RoBERTa & 0.75 & 0.38 & 0.32 & 0.32 \\ \midrule
\multicolumn{6}{c}{\textbf{Multimodality/Fusion}} \\ \midrule
\multicolumn{6}{l}{\textbf{BLOCK fusion}} \\ 
4 & Entity + Text & 0.74 & 0.44 & 0.43 & 0.43 \\
5 & Entity + Image & 0.74 & 0.39 & 0.39 & 0.39 \\
6 & Entity + (Text + Image) & 0.75 & 0.43 & 0.42 & 0.41 \\ \midrule
\multicolumn{6}{l}{\textbf{Attention}} \\ 
7 & Entity + Text & 0.72 & 0.42 & 0.48 & 0.44 \\
8 & Entity + Image & 0.71 & 0.42 & 0.48 & 0.44 \\
9 & Entity + (Text + Image) & 0.71 & 0.42 & 0.49 & 0.44 \\\midrule
\multicolumn{6}{l}{\textbf{Augmentation}} \\ 
10 & Entity + Text (WordNet aug) & 0.76 & 0.48 & 0.46 & \textbf{0.46} \\
11 & Entity + Text (BERT aug) & 0.74 & 0.46 & 0.46 & \textbf{0.46} \\
12 & Entity + Text (Mix aug) & 0.77 & 0.49 & 0.41 & 0.43 \\ \bottomrule
\end{tabular}
}
\caption{Evaluation results on the test set for our classification reformulation of the problem. Our official submission for the shared task is shown in \emph{italic}.}
\label{tab:classification_results}
\end{table}

\begin{table*}[h]
\centering
\setlength{\tabcolsep}{4.0pt}
\scalebox{0.92}{
\begin{tabular}{lrrrrrr|rrrrrr}
\toprule

\multicolumn{1}{c}{\textbf{}} &
\multicolumn{3}{c}{\textbf{E+I, w/o Att.}}  & \multicolumn{3}{c|}{\textbf{E+I, w/ Att.}} & \multicolumn{3}{c}{\textbf{E+[I+T], w/o Att.}}
& \multicolumn{3}{c}{\textbf{E+[I+T], w/ Att.}}\\ \midrule
 \multicolumn{1}{c}{\textbf{Role}} & \multicolumn{1}{c}{\textbf{P}} & \multicolumn{1}{c}{\textbf{R}} & \multicolumn{1}{c}{\textbf{F1}} & \multicolumn{1}{c}{\textbf{P}} & \multicolumn{1}{c}{\textbf{R}} & \multicolumn{1}{c|}{\textbf{F1}} & \multicolumn{1}{c}{\textbf{P}} & \multicolumn{1}{c}{\textbf{R}} & \multicolumn{1}{c}{\textbf{F1}} & \multicolumn{1}{c}{\textbf{P}} & \multicolumn{1}{c}{\textbf{R}} & \multicolumn{1}{c}{\textbf{F1}} \\ \midrule
Hero & 0.06 &	0.02 &	0.03 &	0.09 &	0.15 &	\textbf{0.12} &	0.22 &	0.12 &	0.15 &	0.09 &	0.21 &	0.12 \\
Villain & 0.35 &	0.44 &	0.39 &	0.40 &	0.51 &	\textbf{0.45} &	0.39 &	0.54 &	0.45 &	0.39 &	0.54 &	0.45 \\
Victim &  0.30 &	0.25 &	0.28 &	0.33 &	0.39	& \textbf{0.35} &	0.23 &	0.18 &	0.20 &	0.31 &	0.45 &	\textbf{0.36} \\
Other &  0.86 &	0.84 &	0.85 &	0.88 &	0.81	& 0.84 & 0.87 &	0.84 &	0.85 &	0.89 &	0.77 &	0.82 \\
\bottomrule
\end{tabular}
}
\caption{Role-level results on the test set with (w/) or without (w/o) attention between the context (text, image) and the entity. (E: Entity, I: Image, Att.: Attention, T: Text)}
\label{tab:attention-result}
\end{table*}

\begin{table*}[h]
\centering
\setlength{\tabcolsep}{4.0pt}
\scalebox{0.95}{
\begin{tabular}{lrrrrrrrrr}
\toprule
\multicolumn{1}{c}{\textbf{}} &
\multicolumn{3}{c}{\textbf{No Aug.}}  & \multicolumn{3}{c}{\textbf{Aug. WordNet}} & \multicolumn{3}{c}{\textbf{Aug. BERT}} \\ \midrule
 \multicolumn{1}{c}{\textbf{Role}} & \multicolumn{1}{c}{\textbf{P}} & \multicolumn{1}{c}{\textbf{R}} & \multicolumn{1}{c}{\textbf{F1}} & \multicolumn{1}{c}{\textbf{P}} & \multicolumn{1}{c}{\textbf{R}} & \multicolumn{1}{c}{\textbf{F1}} & \multicolumn{1}{c}{\textbf{P}} & \multicolumn{1}{c}{\textbf{R}} & \multicolumn{1}{c}{\textbf{F1}} \\ \midrule
Hero & 0.21 &	0.12 &	0.15 &	0.33 &	0.21 &	\textbf{0.26} &	0.30 &	0.25 &	\textbf{0.27} \\
Villain & 0.36 &	0.49 &	0.42 &	0.41 &	0.52 &	\textbf{0.46} &	0.39 &	0.51 &	\textbf{0.44} \\
Victim & 0.31 &	0.27 &	0.29 &	0.30 &	0.27 &	0.29 &	0.29 &	0.27 &	0.28 \\
Other & 0.87 & 0.83 &	0.85 &	0.87 &	0.84 &	0.86 &	0.87 &	0.83 &	0.85 \\
\bottomrule
\end{tabular}
}
\caption{Role-level results on the test set for the entity + text combination with and without augmentation.}
\label{tab:augmentation-result}
\end{table*}

\subsection{Role-Level Analysis} 

Next, we studied the impact of using attention and data augmentation on the individual entity roles: \emph{hero}, \emph{villain}, \emph{victim}, and \emph{other}.

Table~\ref{tab:attention-result} shows the impact of using attention on (a)~entity + image (left side), and (b)~entity + [image + text] (right side) combinations. We can observe a sizable gain for the \emph{hero} (+0.09), the \emph{villain} (+0.06), and the \emph{victim} (+0.07) roles in the former case (a). However, for case (b), there is an improvement for the \textit{victim} role only; yet, this improvement is quite sizable: +0.16.

Table~\ref{tab:augmentation-result} shows the impact of data augmentation using WordNet or BERT on the individual roles. We can observe sizable performance gains of +0.11 for the \emph{hero} role, and +0.04 for the \emph{villain} role, when using WordNet-based data augmentation. Similarly, BERT-based data augmentation yields +0.12 for the \emph{hero} role, and +0.02 for the \emph{villain} role. However, the impact of either augmentation on the \emph{victim} and on the \emph{other} role is negligible.

\subsection{Official Submission} 

For our official submission for the task, we used the image modality system from line 1 in Table~\ref{tab:classification_results}, which was quite weak, with a macro-F1 score of 0.23. Our subsequent experiments and analysis pointed to several promising directions: (\emph{i})~combining the textual and the image modalities, (\emph{ii})~using attention, (\emph{iii})~performing data augmentation. As a result, we managed to improve our results to 0.46. Yet, this is still far behind the F1-score of the winning system: 0.5867.

%% file: sections/conclusions.tex
\section{Conclusion and Future Work}
\label{sec:conclusion}

We addressed the problem of understanding the role of the entities in harmful memes, as part of the CONSTRAINT-2022 shared task. We presented a comparative analysis of the importance of different modalities: the text and the image. We further experimented with two task reformulations ---sequence labeling and classification---, and we found the latter to work better. Overall, we obtained improvements when using BLOCK fusion, attention between the image and the text representations, and data augmentation.

In future work, we plan to combine the sequence and the classification formulations in a joint multimodal setting. We further want to experiment with multi-task learning using other meme analysis tasks and datasets. Last but not least, we plan to develop better data augmentation techniques to improve the performance on the low-frequency roles.

\section*{Acknowledgments}
The work is part of the Tanbih mega-project, which is developed at the Qatar Computing Research Institute, HBKU, and aims to limit the impact of ``fake news,'' propaganda, and media bias by making users aware of what they are reading, thus promoting media literacy and critical thinking.